\documentclass[10pt,twocolumn,letterpaper]{article}

\usepackage[final]{cvpr}      % To produce the REVIEW version
\makeatletter
\let\@LN\relax
\makeatother

\usepackage{algorithm}
\usepackage{algorithmic}
\newtheorem{definition}{Definition}
\usepackage{array}
\usepackage{indentfirst}
\usepackage{multirow}
\usepackage{placeins}
\usepackage{etoolbox}
\pretocmd{\linenumbers}{}{}{}

\definecolor{cvprblue}{rgb}{0.21,0.49,0.74}
\usepackage[pagebackref,breaklinks,colorlinks,allcolors=cvprblue]{hyperref}

\def\paperID{27200} % *** Enter the Paper ID here 
\def\confName{CVPR}
\def\confYear{2026}

\title{Dynamic Granularity Matters: Rethinking Vision Transformers Beyond Fixed Patch Splitting}

\author{
	Qiyang Yu\textsuperscript{1},
	Yu Fang\textsuperscript{1,2},
	Yan Chen\textsuperscript{1},
	Jianghao Li\textsuperscript{1},
	Fan Min\textsuperscript{1},
	Tianrui Li\textsuperscript{2},
	Xuemei Cao\textsuperscript{3} \\
	\textsuperscript{1}School of Computer Science and Software Engineering, Southwest Petroleum University \\
	\textsuperscript{2}School of Computing and Artificial Intelligence, Southwest Jiaotong University \\
	\textsuperscript{3}School of Computing and Artificial Intelligence, Southwestern University of Finance and Economics \\
}

\begin{document}
\maketitle
\begin{abstract}
Vision Transformers (ViTs) have demonstrated strong capabilities in capturing global dependencies but often struggle to efficiently represent fine-grained local details. Existing multi-scale approaches alleviate this issue by integrating hierarchical or hybrid features; however, they rely on fixed patch sizes and introduce redundant computation. To address these limitations, we propose Granularity-driven Vision Transformer (Grc-ViT), a dynamic coarse-to-fine framework that adaptively adjusts visual granularity based on image complexity. It comprises two key stages: (1) Coarse Granularity Evaluation module, which assesses visual complexity using edge density, entropy, and frequency-domain cues to estimate suitable patch and window sizes; (2) Fine-grained Refinement module, which refines attention computation according to the selected granularity, enabling efficient and precise feature learning.   Two learnable parameters, $\alpha$ and $\beta$  , are optimized end-to-end to balance global reasoning and local perception. Comprehensive evaluations demonstrate that Grc-ViT enhances fine-grained discrimination while achieving a superior trade-off between accuracy and computational efficiency.
\end{abstract}    
\section{Introduction}

In recent years, the development of deep learning has significantly advanced research in the field of image processing. The proposal of convolutional neural networks have greatly facilitated image classification. From the earliest LeNet \cite{lecun1998gradient} and AlexNet \cite{krizhevsky2017imagenet} to the widely used ResNet \cite{he2016deep}, these works are indispensable contributions to the field. As researchers continue to innovate, the emergence of large-scale models has opened up new opportunities for the field. The Transformer, introduced by Vaswani \cite{vaswani2017attention} as a new attention-based building block for machine translation captures long-range dependencies in sequences. In subsequent research work, the Transformer was adapted to computer vision as the Vision Transformer (ViT)~\cite{dosovitskiy2020image}. ViT preserves the structure of the original Transformer as much as possible, only adjusting the image structure to fit the input section. Experimental results demonstrate that, despite its relatively simple architecture, its performance on public datasets is still powerful and is competitive with traditional CNNs.

\begin{figure}[!t]
	\centering
	\small
	\includegraphics[width=3.0in]{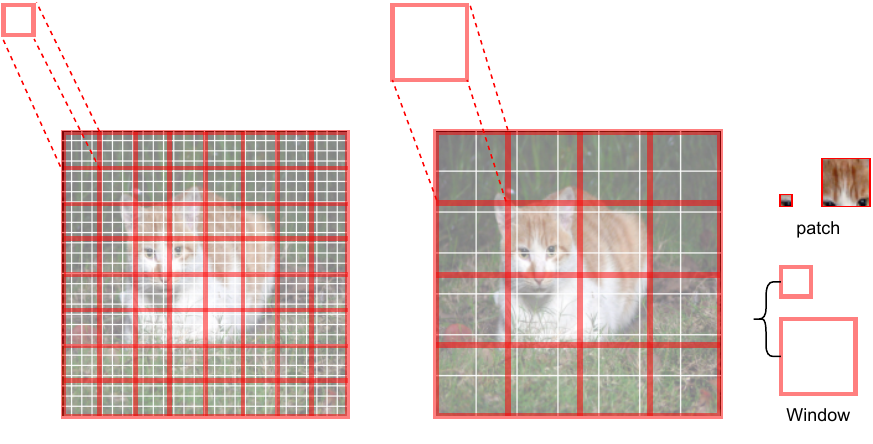}
	\caption{Multi-sized windows in Grc-ViT, with fine-grained windows containing more fine-grained details (left) and coarse-grained windows designed to extract global features (right). Compared to Swin Transformer, which uses a single fixed-size window, Grc-ViT enables more combinations of perceptual regions as windows and patches are varied.}
	\label{f1}
\end{figure} 
 
It has been noted that ViTs inherently lack some of the inductive biases inherent to CNNs \cite{steiner2021train}, such as translation equivariance and spatial locality. Therefore, they often generalize poorly when training on insufficient data. In a nutshell, CNNs rely on local convolutional operations to capture local features of an image and gradually acquire global information as the depth of the network increases. In contrast, Transformers excel to handle long-range dependencies in images by capturing global features directly through a self-attention mechanism. In addition, the transformer was originally designed for text sequences, and simply treating the image data as a one-dimensional vector would ignore its two-dimensional nature. The processing of CNNs is more step-by-step, whereas Transformers appear to achieve success all at once. It is tempting to speculate, that it is the lack of control over local features that may be why Vision Transformers (ViT) require pre-training on a large number of datasets to demonstrate its capabilities. In order for ViT to acquire this capability and improve the generalization of the model. In other words, solving the problem of lack of multi-scale feature extraction ability is the focus of subsequent research.  Research has shown that the inability of ViT to extract multi-scale features is due to the lack of a stage-wise convolutional structure similar to that of convolutional neural networks \cite{liu2021swin}. Multi-scale refers to capturing features at different spatial scales in an image by processing at different resolutions, and utilizing convolution kernels of different sizes to extract low-level detail features and high-level semantic features \cite{lin2017feature}. 

Merely coupling CNNs with ViTs does not fundamentally address this discrepancy, because the model still lacks an explicit mechanism to adjust its feature granularity according to image complexity. This motivates us to revisit how granularity should be selected rather than merely fixed adapted implicitly inside the network. Recent research has explored employing variable patch sizes or multi-granularity segmentation techniques within visual transformers. FlexiViT~\cite{beyer2023flexivit} trains a single model capable of handling multiple patch sizes, while dynamic window transformers~\cite{ren2022beyond} and dynamic transformers~\cite{song2021dynamic} enhance efficiency through adaptive pruning or adjusting segmentation sizes. The recently proposed NaViT (Patch n’Pack) \cite{dehghani2023patch} achieves flexible input resolution, while multi-granularity designs such as CF-ViT \cite{chen2023cf} and MG-ViT \cite{zhang2023mg} enhance model compactness and robustness through hierarchical concatenation strategies. However, most approaches embed patch size selection within the network architecture, achieving this through design schemes such as hierarchical patch embedding, label pruning strategies, or flexible window selection after feature extraction. Crucially, these methods neither explicitly model image complexity nor systematically investigate the relationship between granularity and patch selection.

In this work, we propose the Grc-ViT for image classification. Grc-ViT is based on granularity computation theory \cite{fang2020granularity,yao2018three,yao2013granular} and combines parts of the structure of the swin transformer \cite{liu2021swin}. From the perspective of granularity computing, this approach can enhance the representational capacity of the model without changing the original internal structure of attention calculation. It also highlights the significance of granularity computation for model improvement while ensuring the powerful reasoning and computation capability of ViT. Our work introduces two fundamental innovations upon the existing literature. Firstly, we developed a lightweight, training-free, and convolution-free complexity estimator that integrates spatial, statistical, and frequency-domain descriptors to predict the most suitable granularity prior to any attention computation. Secondly, we design a granularity-adaptive shared Transformer backbone equipped with granularity-level input-output adapters. This enables image patches of varying sizes to share the same attention core while preserving inductive biases specific to each granularity. This design decouples 'complexity inference' from 'feature extraction granularity', offering a novel perspective for multi-granularity visual models. Compared to prior dynamic concatenation approaches, it enhances interpretability, improves routing stability, and refines recognition capabilities.

The contributions of this paper are as follows:
 
\begin{enumerate}
	\item Multi-scale feature extraction is reformulated from a granularity-computing perspective, offering an alternative to CNN-style hierarchies for jointly capturing global context and local details.
	
	\item A lightweight and training-free complexity estimator is introduced to determine the appropriate patch granularity before any attention computation, effectively avoiding the prohibitive cost of fixed small-patch ViTs.
	
	\item To enable collaboration across granularities, a shared transformer backbone is equipped with granularity-specific input and output adapters, which maintain distinct inductive biases while allowing all granularities to benefit from a unified attention core.
	
	\item Comprehensive experiments across multiple fine-grained benchmarks verify that the resulting Grc-ViT achieves higher accuracy with lower inference FLOPs, demonstrating the effectiveness and interpretability of granularity-driven modelling.
\end{enumerate}

\section{Related Works}
\label{sec:formatting}

\textbf{Granular Computing Theoretical Foundation.} Granular computing is an emerging multidisciplinary field that simulates human thinking and addresses complex problems in computational intelligence,  integrating theories and methods from Pawlak's rough set theory \cite{pawlak1982rough} and Zadeh's fuzzy set theory \cite{zadeh1965fuzzy,zadeh1996fuzzy}. It emphasizes the representation and processing of information at various levels of abstraction \cite{fang2022hypersphere}, enabling flexible handling of complex data. Recent research has focused on effectively capturing multi-granular information \cite{chen2024fet,li2024enhancing}. Pedrycz's early work \cite{pedrycz2007granular} highlighted the synergy between granular computing and machine learning, enhancing interpretability in intelligent systems. Applications include data analysis \cite{pedrycz2018granular}, decision support systems \cite{yao2018three,yao2013granular}, and integrating granular computing into deep learning models to improve robustness and interpretability \cite{liu2018granular}. Techniques such as image granulation for facial recognition \cite{Li2016Sequentia}, multi-granularity feature extraction in autoencoders \cite{Zhang2020Sequential}, and granular frameworks for segmentation \cite{pinheiro2024granular} further illustrate its utility. Butenkov \cite{butenkov2004granular} introduced a new granulation technique related to digital image processing and visual understanding, and outlined future applications of this image granulation and processing technique. Additionally, multi-view data fusion, as seen in radar and camera integration for autonomous driving \cite{yao2023radar}, showcases its potential. 

\textbf{Vision Transformer Theoretical Foundation.} CNNs excel at capturing local details and basic information but tend to overlook global context, which is a key focus of the ViT architecture. In contrast, the self-attention mechanism in ViT computes relationships between all input vectors, not just those in the local neighborhood.Through dot products and weighted summation, the self-attention mechanism is able to capture global contextual information. ViT leverages a self-attention mechanism to capture global dependencies without assuming local structures, enabling it to learn richer feature representations when sufficient data and computational resources are available. However, the lack of inductive biases present in CNNs can be a limitation, as ViT divides images into fixed-size patches, which may overlook fine-grained details and local context \cite{heo2021rethinking}. Originally designed for text sequences, ViT treats image data as one-dimensional vectors, ignoring their two-dimensional nature. To address these limitations, HaViT \cite{li2022havit} introduces a hybrid attention mechanism for patch relationships, PEDTrans \cite{lin2022pedtrans} proposes a Patch Enhancement Module and a Dropout Patch Module for small image regions, and FlexVIT \cite{beyer2023flexivit} explores the impact of patch size on ViT's modeling capabilities, laying the groundwork for integrating granular computing with ViT.

\section{Methodology}
\subsection{Formulation}
In Section 3, we provide preliminary theoretical foundations and a comprehensive description of our model architecture. ViT divides the image into a sequence of patches, transforming the image data into tokenized words that are compatible with the transformer's input format \cite{dosovitskiy2020image}. We firmly believe that the choice of patch size is critical for effective feature extraction, as demonstrated in the subsection \ref{sec_patch}. Different patch sizes capture image features at varying granularities, with the patch size determining the level of detail or abstraction of the extracted features. Smaller patches emphasize fine-grained details, while larger patches focus on more global features of the image. Since images often contain spatially heterogeneous content with varying features across different regions, incorporating patches with different granularities enables the model to better adapt to the spatial variability inherent in real-world scenarios. This approach offers a promising solution to the problem of ViT's lack of inductive bias.
\begin{figure}[!htpp]
	\centering
	\small
	\subfloat[]{\includegraphics[width=1.5in]{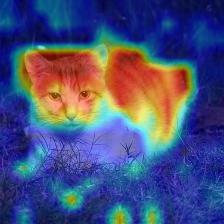}%
		\label{f2_a}}
	\hfil
	\subfloat[]{\includegraphics[width=1.5in]{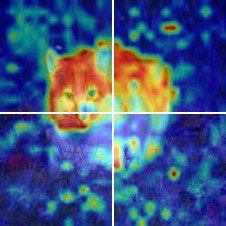}%
		\label{f2_b}}
	\hfil
	\subfloat[]{\includegraphics[width=1.5in]{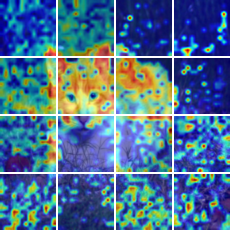}%
		\label{f2_c}}
	\hfil
	\subfloat[]{\includegraphics[width=1.5in]{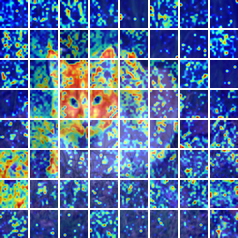}%
		\label{f2_d}}
	\caption{Effect of granularity on feature extraction. (a) shows the hotspot map of the attention mechanism on the original picture. Respectively, (b), (c), and (d) show the hotspot maps of the attention mechanism on the feature maps at different granularity levels.}
	\label{f2}
\end{figure}

Fig. \ref{f2} visualizes the attention mechanism across different granularity levels, illustrating how attention is focused on feature maps at varying granularities within a multi-granularity framework. The original image depicts a cat, with no processing applied to it at this stage. We observed that certain parts of the background, which are connected to the main object, were also identified as focal points. As the granularity level progresses from coarse to fine, the patch divisions become smaller. The cat's head and body are highlighted, with the attention mechanism focusing more closely on the main features. Smaller patches capture finer local details and reveal more specific information. At the finest granularity level, the majority of the attention is concentrated on the facial region, which is considered the most important feature of the image. For instance, in a classification task, the categories may include not only 'cat' but also 'dog'. For a machine, both categories share similar features, except for the head. Thus, to correctly classify these two categories, extracting facial features becomes crucial. However, if the patch is too small, the model may be distracted by background noise, affecting its focus. Therefore, we combine granularity levels with patches, where granularity determines patch selection, and patches guide granularity, creating a nested structure that forms a granularity-driven visual model.

\begin{figure*}[!t]
	\centering
	\small
	\includegraphics[width=\textwidth]{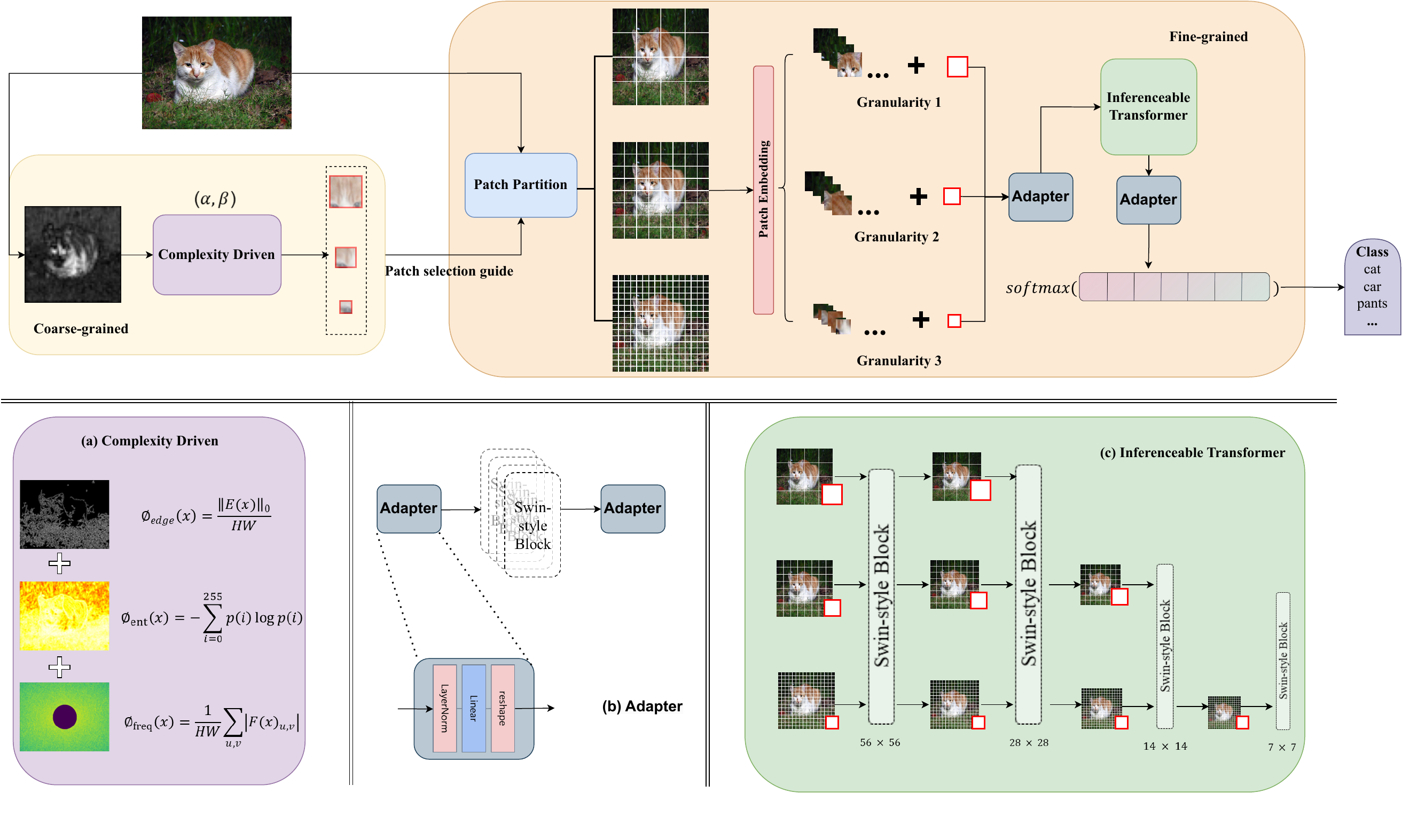}
	\caption{Structural diagram of the model of Grc-ViT. Grc-ViT optimizes feature extraction through a dual-granularity synergistic mechanism, where coarse-graining guides fine-graining cascading while taking into account efficiency and accuracy, and verifies the appropriateness of local information distribution and granularity selection. Coarse granularity globally filters global key regions, quickly locates potential semantic units, and reduces computational redundancy. Fine-grained based on the results of coarse-grained, dynamic selection of multi-level granularity, gradually refine local features.}
	\label{f3}
	\hfil
\end{figure*}

Grc-ViT consists of two main components: coarse-grained feature extraction and fine-grained attention computation. The coarse level provides a kind of summary description for the fine level, while the fine level provides more detailed information for the coarse level. Fig. \ref{f3} indicates that an overview of the Grc-ViT architecture. In the coarse-grained stage, preliminary features are extracted from the input. These initial features are then analyzed for their feature complexity, which is compared against learnable thresholds to guide the selection of patch sizes for the fine-grained stage. Feature complexity plays a crucial role in determining the appropriate granularity for the input, effectively guiding the patch size for dividing the image at the fine-grained level. At the fine-grained stage, attention computation and feature map extraction are performed. The patch size used for attention computation is derived from the coarse-grained stage. Different patch sizes correspond to varying granularity levels, where each level determines the fineness of the corresponding features. Coarser granularity levels result in less detailed features, whereas finer granularity levels yield more detailed features. In essence, the structure of the model at the fine-grained stage is dynamically determined by the granularity level. The feature computation module is designed to adapt to the granularity of the patches, ensuring that the level of detail aligns with the input's complexity.
\subsection{Coarse-grained Complexity Extraction}

The objective of the coarse-grained stage is to estimate the visual complexity of an input image and determine an appropriate granularity level for subsequent fine-grained attention computation. Instead of adopting fixed thresholds or heuristic rules, we introduce a differentiable complexity-driven mechanism that maps each image to a granularity level in a data-dependent manner.

Let $x \in \mathbb{R}^{H \times W \times 3}$ denote an input image. 
We characterize its visual complexity using three lightweight and complementary cues:

\begin{itemize}
	\item \textbf{Edge density.} 
	Applying the Canny operator produces an edge map $E(x)$. Measure the scale of structural boundaries to capture features within the spatial domain.
	The normalized edge density is denoted as
	\begin{equation}
		\phi_{\mathrm{edge}}(x) = \frac{\| E(x) \|_0}{H W},
	\end{equation}
	where $\| E(x) \|_0$ counts nonzero edge pixels.
	
	\item \textbf{Information entropy.}
	 The randomness describing intensity distributions captures information within the field of statistics. Let $p(i)$ be the empirical distribution of pixel intensities.
	The Shannon entropy is computed as
	\begin{equation}
		\phi_{\mathrm{ent}}(x)
		= - \sum_{i=0}^{255} p(i)\log p(i),
	\end{equation}
	
	\item \textbf{Frequency-domain variation.}
	Measure the proportion of high-frequency energy, capturing the intricate complexity of fine details in the frequency domain. Let $\mathcal{F}(x)$ denote the magnitude of the 2D Fourier transform.
	The normalized amplitude variation is given by
	\begin{equation}
		\phi_{\mathrm{freq}}(x)
		= \frac{1}{HW}\sum_{u,v} \big| \mathcal{F}(x)_{u,v} \big|.
	\end{equation}
\end{itemize}

The three cues are normalized and fused into a single differentiable complexity score:
\begin{equation}
	\Phi(x)
	= w_1 \phi_{\mathrm{edge}}(x)
	+ w_2 \phi_{\mathrm{ent}}(x)
	+ w_3 \phi_{\mathrm{freq}}(x),
\end{equation}
where $w_1, w_2, w_3$ are learnable parameters optimized jointly with the coarse-stage. These three views jointly provide a robust, model-free description of image complexity, aligning naturally with the granular computing principle: simple images require coarse granularity, while detail-rich images demand fine granularity.

Following the principle of multi-granularity analysis in granular computing, 
we introduce two learnable thresholds $\alpha$ and $\beta$ to partition the complexity axis. 
To ensure valid ordering, they are parameterized as
\begin{equation}
	\alpha = \sigma(a), \qquad 
	\beta = \alpha + (1-\alpha)\sigma(b),
\end{equation}
where $a,b \in \mathbb{R}$ are unconstrained trainable variables and $\sigma(\cdot)$ denotes the sigmoid function.

Thus, the granularity level assigned to image $x$ is
\begin{equation}
	s(x) = 
	\begin{cases}
		1, & \Phi(x) < \alpha, \\[2pt]
		2, & \alpha \le \Phi(x) < \beta, \\[2pt]
		3, & \Phi(x) \ge \beta.
	\end{cases}
\end{equation}

The coarse stage outputs a discrete granularity label $s(x)\in\{1,2,3\}$. This label determines the corresponding patch--window configuration used in the fine-grained attention stage. 
Notably, all parameters $(w_1,w_2,w_3, \alpha, \beta)$ are optimized end-to-end, 
making granularity selection fully data-driven and differentiable.
\subsection{Fine-grained Attention Computation} 
In the fine-grained stage, we adopt the Swin transformer \cite{liu2021swin} as the backbone network for attention computation. The Swin Transformer constructs a hierarchical network structure, where attention computation is restricted to non-overlapping windows within the input frame. Specifically, the model employs a patch size of 4 and a window size of 7. While retaining the multi-head attention mechanism, the introduction of the windowing strategy significantly reduces computational complexity compared to global attention. For Grc-ViT, we revisit the relationship between patches and windows, redefining patches in terms of granularity. In our approach, the patch size is dynamically determined by the current granularity level, and the window size is no longer fixed. This design introduces a flexible model structure, transforming the architecture into a sequential inference model that adapts to varying granularities. Based on this, we propose def. \ref{def_1} to formalize our approach.
\begin{definition}
	{\itshape For the domain $U$, Let Grc = \{$Grc_1$, $Grc_2$,..., $Grc_g$\} be set the g-levels of the granularity, where $Grc_g$=\{$U_j$, $patch\_size_g$, $window\_size_g$, g$\mid$g = 1, 2,..., g\}, $g$ denotes the granularity level, $U_j$ denotes the decision object, $patch\_size_g$ and  $window\_size_g$ are the attributes of the decision object.}
	\label{def_1}
\end{definition}

According to def. \ref{def_1}, with respect to the object of this paper, we provide three hierarchical reasoning structures with different granularities at the fine-grained level for models to choose from. Therefore, we propose a more specific definition \ref{def_2}.
\begin{definition}
	{\itshape Let Grc = \{$Grc_1$, $Grc_2$, $Grc_3$\}, where $Grc_1$ = \{$p_1$=16, $w_1$=28\}, $Grc_2$ = \{$p_2$=8, $w_2$=14\}, and $Grc_3$ = \{$p_3$=4, $w_3$=7\}.}
	\label{def_2}
\end{definition}
Def. \ref{def_2} highlights two key parameters patch\_size $p_g$ and window\_size $w_g$ as decision condition attributes in the inference model structure.

While the coarse stage determines the appropriate granularity (\textbf{coarse(1)}, \textbf{medium(2)}, or \textbf{fine(3)}) for each input image, the fine stage performs classification using a granularity-specific transformer pipeline. Despite the heterogeneous patch sizes and sequence lengths, all granularities share the same attention core. This is made possible by our proposed granularity adapters and a unified shared attention core. Together they enable multi-granularity computation while keeping the overall model lightweight.

\textbf{Dynamic patch embedding}
For granularity $g\in\{1, 2, 3\}$ corresponding to patch sizes
$p_g\in\{16,8,4\}$, the image is first split into patches and projected into
a granularity-specific embedding dimension $D_g$:
\begin{equation}
	\mathbf{X}^{(g)}_0
	= \phi_g(\mathbf{I})
	\in \mathbb{R}^{N_g\times D_g},
\end{equation}
where $N_g=(H/p_g)(W/p_g)$.
This module is implemented by Dynamic PatchEmbed, which mirrors Swin’s patchification but with granularity-specific projection layers.

\textbf{Granularity-specific input adapter}
Because different granularities produce different embedding dimensions, we introduce an input adapter
$\psi^{\mathrm{in}}_g:\mathbb{R}^{D_g}\rightarrow\mathbb{R}^{D_{\mathrm{uni}}}$:
\begin{equation}
	\hat{\mathbf{X}}_0
	= \psi^{\mathrm{in}}_g(\mathbf{X}^{(g)}_0)
	\in \mathbb{R}^{N_g\times D_{\mathrm{uni}}},
\end{equation}
which maps heterogeneous token spaces into a unified representation dimension $D_{\mathrm{uni}}$. This makes the downstream transformer blocks fully shared across granularities.

\textbf{Shared attention core}
The unified tokens $\hat{\mathbf{X}}_0$ are then processed by a sequence of granularity-agnostic transformer blocks:
\begin{equation}
	\hat{\mathbf{X}}_{\ell+1}
	= \mathcal{B}_\ell(\hat{\mathbf{X}}_{\ell}),
	\qquad \ell=0,\ldots,L_g-1,
\end{equation}
where $L_g$ is the number of layers assigned to granularity $g$. Each block $\mathcal{B}_\ell$ consists of:
\begin{itemize}
	\item \textbf{Window-based multi-head self-attention} implemented via a
	unified Shared Attention, shared across all granularities.
	\item \textbf{Cyclic shift} of window partitions depending on the layer index:
	even layers use regular windows while odd layers use shifted windows,
	following Swin’s design.
	\item \textbf{Shared MLP} for token-wise feed-forward transformation.
\end{itemize}

Let $\mathcal{W}(\cdot)$ denote window partitioning and $\mathrm{Attn}$ the shared attention kernel. Given window size $w_g\in\{28,14,7\}$, we compute:
\begin{equation}
	\mathbf{X}_{\text{win}}
	= \mathcal{W}(\hat{\mathbf{X}}_{\ell}),
\end{equation}
\begin{equation}
	\mathbf{X}_{\text{att}}
	= \mathrm{Attn}(\mathbf{X}_{\text{win}}),
\end{equation}
and reverse window partitioning to restore spatial structure.

\textbf{Granularity-specific output adapter}
After processing by shared blocks, we obtain:
\begin{equation}
	\hat{\mathbf{Z}}
	= \hat{\mathbf{X}}_{L_g}
	\in \mathbb{R}^{N_g\times D_{\mathrm{uni}}}.
\end{equation}
Before classification, we apply the output adapter
$\psi^{\mathrm{out}}_g:\mathbb{R}^{D_{\mathrm{uni}}}\rightarrow\mathbb{R}^{D_g}$:
\begin{equation}
	\mathbf{Z}^{(g)}
	= \psi^{\mathrm{out}}_g(\hat{\mathbf{Z}})
	\in \mathbb{R}^{N_g\times D_g},
\end{equation}
which restores the granularity-specific semantic space.
This ensures that different granularities maintain distinct inductive
biases while benefiting from the shared attention core.

Swin Transformer \cite{liu2021swin} provides an analysis of the computational complexity of the model. The computational complexity of Swin Transforemr is $\colon$

\begin{equation}
	\Omega(\mathrm{Swin}) = \sum_{l=1}^{L}(4H_lW_lC^2+2H_lW_lC(M)^2),
	\label{eq3}
\end{equation}
From Eq. \ref{eq3}, we get the computational complexity of Grc-ViT is $\colon$ 

\begin{equation}
	\Omega(\mathrm{Grc-ViT}) = \sum_{l=1}^{L}(3H_lW_lC^2+2H_lW_lC(M_l)^2+3H_lW_lC),
	\label{eq4}
\end{equation}where $M$ is window\_size, $H_l$ and $W_l$ are the height and width of the image, $C$ is the number of channels in the image and  $l$ is the number of layers in the model. 

The optimization of Grc-ViT is reflected in: dynamic window adjustment and calculation volume allocation. Grc-ViT uses a smaller $M_l$ at depth, which greatly reduces the cumulative value of $(M_l)^2$. Meanwhile, Grc-ViT captures global features with a large window at shallow layers (higher computation) and refines local details with a small window at deeper layers (significantly lower computation), which results in better overall complexity.
\section{Experiments}
In Section 4, We present the experimental part of the paper, which includes experimental preparation, multi-granularity analysis, ablation and comparison experiments.

\subsection{Experimental Preparations}
\textbf{Experimental device} All models were trained on a server equipped with a single NVIDIA 3090 graphics card, while all evaluation tasks were performed on a native NVIDIA 3060 graphics card.
\textbf{Datasets} To ensure the credibility of the evaluation data, and considering the limited computational power of the experimental setup, we selected three foundational public datasets: CIFAR-10\footnote{\footnotesize CIFAR: \url{https://www.cs.toronto.edu/~kriz/cifar.html}}, CIFAR-100\footnotemark[1], and Tiny-ImageNet\footnote{\footnotesize Tiny-ImageNet: \url{http://cs231n.stanford.edu/tiny-imagenet-200.zip}}. Based on the capacity of the dataset and the number of categories, we consider that the complexity of the dataset increases in a gradient from easy to difficult. To further demonstrate the effectiveness of our model in capturing 
local subtle features and discriminative fine-grained structures, we additionally adopt three classical fine-grained benchmarks: FGVC Aircraft\footnote{\footnotesize FGVC Aircraft:\url{https://www.robots.ox.ac.uk/~vgg/data/fgvc-aircraft/}}, Stanford Dogs\footnote{\footnotesize Stanford Dogs: \url{http://vision.stanford.edu/aditya86/ImageNetDogs/}}, and CUB-200-2011\footnote{\footnotesize CUB-200-2011: \url{http://www.vision.caltech.edu/datasets/cub_200_2011/}}. These datasets contain numerous visually similar subcategories, where category distinctions rely heavily on local textures, part-level cues, and subtle appearance differences. Therefore, they serve as rigorous testbeds for evaluating the model's ability to extract stable, fine-grained features under limited inter-class variation.
\begin{table}[!h]
	\centering
	\caption{Description of the data set.}
	\small
	\begin{tabular}{lcc}
		\toprule
		Dataset & volume & categories  \\  \midrule
		CIFAR-10 & 60,000 & 10 \\
		CIFAR-100 & 60,000 & 100  \\
		Tiny-ImageNet & 110,000 &200  \\
		FGVC Aircraft & 10,000 & 100  \\ 
		Stanford Dogs & 20,580 & 120  \\ 
		CUB-200-2011 & 117,800 & 200  \\ 
		\bottomrule
	\end{tabular}
	\label{tab_2}
\end{table}
\subsection{Parameters Analysis}
\label{4.2}
Table \ref{tab_3} shows the relationship between patch\_size and computational complexity by reasoning about ViT. The patch\_size for the standard ViT modeling process is set to $16\times16$. We note that as the patch size increases, the corresponding Flops decrease. Conversely, as the patch size decreases, the corresponding Flops increase. This is because the complexity of the self-attention layer is proportional to the square of the number of patches N and inversely proportional to the patch\_size. The research work in this paper is dedicated to obtaining multi-granularity features while ensuring that a balance is found between model accuracy and computational cost. The data in Table \ref{tab_3} validates why it is important to set up the model as a multi-granularity reasonable hierarchical model rather than putting all the reasoning processes at the same granularity level.
\begin{table}[!h]
	\caption{Multi-granularity patch analysis.}
	\centering
	\small
	\begin{tabular}{lccc}
		\toprule 
		patch\_size & $32\times32$ & $16\times16$ & $8\times8$\\  \midrule
		Flops&4G&17G&76G\\
		\bottomrule
	\end{tabular}
	\label{tab_3}
\end{table} 

We treat the two granularity thresholds $\alpha$ and $\beta$ as learnable parameters, optimising them jointly with the coarse-grained stage. During training, the model automatically adjusts $(\alpha,\beta)$ to adapt to the complexity distribution of the dataset: $\alpha$ gradually increases, $\beta$ gradually decreases, and the intermediate fuzzy interval $(\alpha,\beta)$ narrows accordingly.
This indicates heightened confidence in the coarse-grained stage to distinguish between low- and high-complexity images,
rendering medium-complexity samples increasingly sparse.
Upon convergence, the learned thresholds stabilise
and are directly employed for granularity routing during inference.
\begin{figure}[!h]
	\includegraphics[width=3.5in]{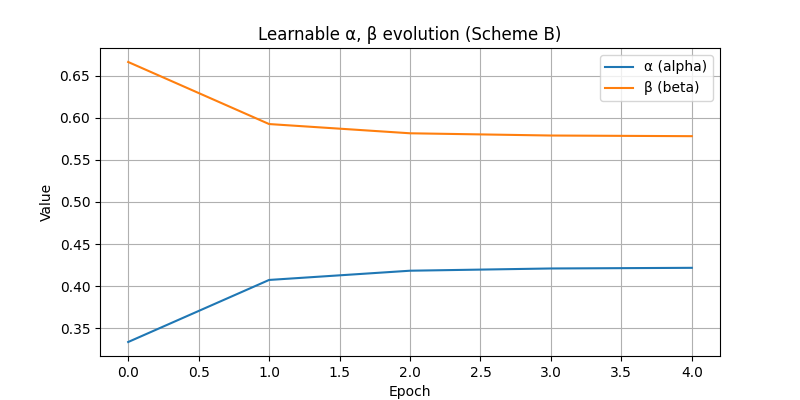}
	\caption{The process of fine-tuning parameters $\alpha$ and $\beta$.}
	\label{fig_4}
\end{figure}

\subsection{Ablation Study}
Table \ref{tab_4} and Table \ref{tab_5} show the ablation experiments of Grc-ViT. The whole Grc-ViT is based on the idea of granularity computation, and W/O stands for no granularity driver module. In both datasets, CIFAR-10 and Tiny-ImageNet, the accuracy of the Grc-ViT model is improved over the baseline model without the granularity driver module. This highlights the effectiveness of Grc-ViT in feature extraction. In the CIFAR-100 dataset, the accuracy of Grc-ViT is slightly lower than the baseline model, which may be due to the instability of the training process. These findings suggest that Grc-ViT consistently improves model performance on a variety of datasets, especially in terms of feature representation and learning capability. The inference time of the Grc-ViT model was reduced by 76.87 seconds compared to 93.84 seconds for the W/O model. The significant reduction in inference time highlights the efficiency of the Grc-ViT architecture, making it more suitable for real-time applications. Furthermore, the effectiveness of the granularity-driven module is corroborated through ablation experiments presented in the section \ref{sec_sup} of supplementary materials.
\begin{table}[!htpp]
	\caption{Classification Ablation Experiment.}
	\centering
	\small
	\begin{tabular}{lm{1.5cm}m{1.5cm}m{1.5cm}}
		\toprule 
		Methods & CIFAR-10 & CIFAR-100 & Tiny-Imagenet\\  \midrule
		W/O&97.87&\textbf{85.86}&79.25\\
		%\hline
		Grc-ViT(Ours)&\textbf{98.54}&85.63&\textbf{79.72}\\
		\bottomrule
	\end{tabular}
	\label{tab_4}
\end{table}
\begin{table}[!h]
	\centering
	\caption{Model Complexity and Inference Time.}
	\small
	\begin{tabular}{lccc}
		\toprule
		Methods & Flops & \#Param. & Inference Time\\  
		\midrule
		W/O&15.14G&86.95M&93.84s\\
		Grc-ViT(Ours)&\textbf{6.07G}&\textbf{61.73M}&\textbf{76.87s}\\
		\bottomrule
	\end{tabular}
	\label{tab_5}
\end{table}

To verify that our method improves local discrimination, we remove the granularity module and compare against the baseline model ('W/O'). Table \ref{tab_6}  indicates Grc-ViT yields consistent gains on FGVC Aircraft and Stanford Dogs—two datasets where high intra-class similarity requires stronger local feature sensitivity. These results confirm that granularity-aware processing enhances fine-grained recognition by enabling the model to capture subtle local cues that plain ViTs often overlook.

\begin{table}[!htpp]
	\centering
	\caption{Localised Improvement Ablation Experiment.}
	\small
	\begin{tabular}{lm{1.5cm}m{1.5cm}m{1.5cm}}
		\toprule 
		Methods & FGVC Aircraft & Stanford Dogs &CUB-200-2011 \\  \midrule
		W/O&79.51&80.46&\textbf{85.23}\\
		%\hline
		Grc-ViT(Ours)&\textbf{81.54}&\textbf{81.63}&83.72\\
		\bottomrule
	\end{tabular}
	\label{tab_6}
\end{table}
\subsection{Comparison Study}
\textbf{Baseline Methods} We selected nine models that are more popular in the current research field as baseline methods: ViT \cite{dosovitskiy2020image}, Efficientformer \cite{li2022efficientformer}, BEiT \cite{bao2021beit}, MobileViT \cite{mehta2021mobilevit}, DaViT \cite{ding2022davit}, MViT \cite{fan2021multiscale}, PoolFormer \cite{yu2022metaformer}, Rev-Vit \cite{mangalam2022reversible}, T2T-ViT \cite{yuan2021tokens}. To achieve a more effective comparison and address the limitations of computing power, we selected baseline models based on their base or small versions. These models utilize parameter weights pre-trained on ImageNet-1K.

\textbf{Evaluation of indicators} To demonstrate the effectiveness of Grc-ViT, we focused on two key aspects to validate the model's performance metrics: ablation experiments and comparative experiments. To evaluate the model's performance across different datasets, we selected four evaluation metrics: \textbf{Accuracy} and \textbf{F1 score}. Additionally, we calculated both the \textbf{Flops} and the number of \textbf{Parameters} for the model.
\begin{table}[!htb]
	\captionsetup[table]{labelfont=bf, textfont=bf, labelsep=newline, justification=raggedright, singlelinecheck=false}
	\small
	\caption{Average Accuracy/F1-score of models.}
	\centering
	\begin{tabular}{@{\extracolsep{\fill}}l ccc@{}}
		\toprule
		\multirow{2}{*}{Model} & 
		\multicolumn{3}{c}{Dataset} \\
		\cmidrule[0.6pt](lr){2-4}
		& CIFAR-10 & CIFAR-100 & Tiny-ImageNet \\
		\cmidrule(lr){2-2} \cmidrule(lr){3-3} \cmidrule(lr){4-4}
		ViT & 95.55/95.55 & 80.50/80.50 & 77.95/77.97 \\
		Efficientformer & 95.09/95.08 & 80.29/80.47 & 78.26/78.26 \\
		BEiT & 96.56/96.56 & 84.68/84.61 & 76.86/76.86 \\
		MobileViT & 92.92/92.90 & 79.86/79.80 & 73.83/73.84 \\
		DaViT & 96.86/96.86 & 84.46/84.54 & 76.92/76.92 \\
		MViT & 96.53/96.53 & 79.39/79.39 & 77.84/77.86 \\
		PoolFormer & 96.75/96.74 & 82.28/82.23 & 76.43/76.40 \\
		Rev-ViT & 97.09/97.09 & 79.00/79.14 & 76.56/76.56 \\
		T2T-ViT & 97.39/97.39 & 83.83/83.75 & 76.50/76.50 \\
		Grc-ViT (Ours) & \textbf{97.54/97.54} & \textbf{85.63/85.46} & \textbf{79.72/79.73} \\
		\bottomrule
	\end{tabular}
	\label{tab_7}
\end{table}

The Grc-ViT model greatly enhances feature extraction by integrating a granularity-based computational module. Results on the CIFAR-10 and CIFAR-100 datasets validate this feature, with Grc-ViT exceeding the accuracy of traditional ViT models. These improvements enable Grc-ViT to better understand and categorize subtle features in complex images, especially in multi-class tasks where it shows greater robustness.
In addition to enhanced feature extraction, Grc-ViT also optimizes computational efficiency. Table \ref{tab_8} indicates that Grc-ViT has 6.07G FLOPS, 61.73M parameters, and 76.87 seconds inference time. Compared with ViT, Grc-ViT effectively utilizes the computational resources. By reducing computational complexity, it not only speeds up inference, but also reduces resource requirements, making it more feasible for real-world applications. In a comparison of ten models, the FLOPs of Grc-ViT ranked fourth, which indicates that the concept of granular computing can simplify the computation to a certain extent, although it has not yet reached the optimal performance. 
\begin{table}[!h]
	\centering
	\caption{Complexity and Inference Time of models.}
	\small
	\begin{tabular}{lrrr}
		\toprule
		Model & Flops & \#Param. & Inference Time\\  
		\midrule
		ViT&16.86G&88.19M&82.56s\\
		%\hline
		Efficientformer&2.6G&12.3M&\textbf{24.96s}\\
		%\hline
		BEiT&12.67G&86M&78.14s\\
		%\hline
		MobileViT&\textbf{1.44G}&\textbf{5.7M}&33.78s\\
		%\hline
		DaViT&15.24G&86.94M&94.75s\\
		%\hline
		MViT&70.5G&36.6M&91.45s\\
		%\hline
		PoolFormer&4.0G&12M&24.42s\\
		%\hline
		Rev-ViT&17.6G&81.8M&77.93s\\
		%\hline
		T2T-ViT&9.6G&21.5M&47.65s\\
		%\hline
		Grc-ViT(Ours)&6.07G&61.73M&76.87s\\
		\bottomrule
	\end{tabular}
	\label{tab_8}
\end{table}

This result shows that Grc-ViT maintains stable performance on datasets of different sizes, demonstrating its strong generalization ability. This adaptability enables Grc-ViT to effectively perform classification tasks even with limited data or unbalanced samples.
This innovation not only improves accuracy, but also optimizes computational efficiency. Grc-ViT establishes a new feature extraction strategy that strikes a balance between global context and local details. This balance makes Grc-ViT more capable in processing complex images, especially in dealing with various visual tasks.
\section{Conlusion}
In conclusion, the Grc-ViT effectively overcomes the limitations of traditional Vision Transformers by enhancing local feature extraction while maintaining global reasoning capabilities. Its innovative two-phase framework dynamically adjusts patch and window sizes based on granularity levels, resulting in superior fine-grained detail recognition. Moreover, the ingenious integration of granular computation principles into computer vision establishes a novel paradigm for related research. Experimental results demonstrate that Grc-ViT improves local feature representation while preserving the global inference strengths of ViTs, positioning it as a promising solution for tasks requiring balanced local and global feature integration in computer vision.
\section{Acknowledgments}
We would like to express our gratitude to the large language model (GPT-5) for its invaluable assistance and refinement during the paper writing process.
{
    \small
    \bibliographystyle{ieeenat_fullname}
    \bibliography{main}
}

\clearpage
\appendix
\renewcommand{\thesection}{\Alph{section}}
\renewcommand{\thesubsection}{\Alph{section}.\arabic{subsection}}

\clearpage
\setcounter{page}{1}
\setcounter{section}{0}
\renewcommand{\thesection}{\Alph{section}}
\renewcommand{\thesubsection}{\thesection.\arabic{subsection}}
\maketitlesupplementary

\section{Algorithm Details}
\begin{algorithm}[htbp]
	\caption{Coarse Stage: Complexity Estimation and Granularity Selection}
	\label{alg:coarse}
	\begin{algorithmic}[1]
		\REQUIRE Image $\mathbf{x} \in \mathbb{R}^{H \times W \times 3}$; learnable weights $\mathbf{w}\in\mathbb{R}^3$; thresholds $(\alpha,\beta)$.
		\ENSURE Complexity score $s \in (0,1)$; granularity index $g\in\{1,2,3\}$.
		
		\STATE $gray \gets \text{Gray}(\mathbf{x})$.
		
		\STATE $c \gets \phi_{\mathrm{canny}}(gray)$.
		\STATE $e \gets \phi_{\mathrm{entropy}}(gray)$.
		\STATE $f \gets \phi_{\mathrm{fft}}(gray)$.
		
		\STATE $\mathbf{v} \gets [c, e, f]$.
		
		\STATE $\tilde{\mathbf{w}} \gets \mathrm{softmax}(\mathbf{w})$.
		
		\STATE $z \gets \tilde{\mathbf{w}}^\top \mathbf{v}$.
		
		\STATE $\Phi(x) \gets \sigma(z)$.
		
		\IF{$\Phi(x) < \alpha$}
		\STATE $g \gets 0$
		\ELSIF{$\Phi(x) < \beta$}
		\STATE $g \gets 1$
		\ELSE
		\STATE $g \gets 2$
		\ENDIF
		
		\RETURN $(\Phi(x), g)$.
	\end{algorithmic}
\end{algorithm}
The coarse stage begins by converting the input RGB image into a grayscale representation, which serves as a unified basis for all subsequent complexity measurements. From this grayscale image, three complementary descriptors are extracted to characterize different aspects of visual complexity. A spatial descriptor is computed using the Canny operator, where the proportion of detected edges reflects the amount of structural detail present in the image. A statistical descriptor is then obtained by computing the normalized Shannon entropy of the grayscale histogram, capturing the distribution of intensity variations. Next, a frequency-domain descriptor is derived by applying a 2D Fourier transform, suppressing the central low-frequency region, and measuring the ratio of high-frequency energy, which represents the amount of fine texture information. These descriptors form a three-dimensional feature vector, which is combined using a softmax-normalized learnable weight vector. The resulting scalar response is passed through a sigmoid function to produce the final complexity score $s \in (0,1)$. This score is compared against two thresholds $(\alpha,\beta)$: images with $s < \alpha$ are assigned coarse granularity, those with $\alpha \le s < \beta$ are assigned medium granularity, and those with $s \ge \beta$ are routed to the fine-grained branch. The coarse stage outputs both the complexity score and the corresponding granularity index.

\begin{algorithm}[t]
	\caption{Fine Stage: Granularity-Adaptive Shared Transformer}
	\label{alg:fine}
	\begin{algorithmic}[1]
		\REQUIRE Granularity index $g \in \{1,2,3\}$;
		granularity-specific patch embedding $\phi^{(g)}_{\mathrm{patch}}$;
		input adapter $\psi^{\mathrm{in}}_g$;
		output adapter $\psi^{\mathrm{out}}_g$;
		shared transformer depth $L$;
		input image $\mathbf{x}$.
		\ENSURE Granularity-specific token representation $\mathbf{Z}^{(g)}$.
		
		\STATE \textbf{Patch embedding (granularity $g$):}
		\STATE \quad $\mathbf{X}^{(g)}_0 \gets \phi^{(g)}_{\mathrm{patch}}(\mathbf{x}) \in \mathbb{R}^{N_g \times D_g}$.
		\STATE \textbf{Input adapter:} project to unified dimension:
		\STATE \quad $\hat{\mathbf{X}}_0 \gets \psi^{\mathrm{in}}_g(\mathbf{X}^{(g)}_0) \in \mathbb{R}^{N_g \times D_{\mathrm{uni}}}$.
		
		\FOR{$\ell = 1$ to $L$}
		\STATE \textbf{LayerNorm before attention:}
		\STATE \quad $\mathbf{U}_{\ell-1} \gets \mathrm{LN}(\hat{\mathbf{X}}_{\ell-1})$.
		\STATE \textbf{Window-based multi-head self-attention:}
		\STATE \quad Partition $\mathbf{U}_{\ell-1}$ into local windows according to $g$;
		\STATE \quad Compute query/key/value in each window:
		\STATE \qquad $\mathbf{Q},\mathbf{K},\mathbf{V} \gets \mathrm{Linear}_{QKV}(\mathbf{U}_{\ell-1})$;
		\STATE \quad Compute window attention with relative position bias:
		\STATE \qquad $\mathrm{Attn} \gets \mathrm{Softmax}\!\big(\frac{\mathbf{Q}\mathbf{K}^\top}{\sqrt{d_h}} + \mathbf{B}_{\mathrm{rel}}\big)\mathbf{V}$;
		\STATE \quad Merge windows and apply output projection:
		\STATE \qquad $\tilde{\mathbf{H}}_\ell \gets \mathrm{Linear}_{\mathrm{out}}(\mathrm{Attn})$.
		\STATE \textbf{Residual connection (attention block):}
		\STATE \quad $\mathbf{H}_\ell \gets \hat{\mathbf{X}}_{\ell-1} + \tilde{\mathbf{H}}_\ell$.
		\STATE \textbf{LayerNorm before MLP:}
		\STATE \quad $\mathbf{V}_\ell \gets \mathrm{LN}(\mathbf{H}_\ell)$.
		\STATE \textbf{Feed-forward network (MLP) with residual:}
		\STATE \quad $\tilde{\mathbf{X}}_\ell \gets \mathrm{MLP}(\mathbf{V}_\ell)$.
		\STATE \quad $\hat{\mathbf{X}}_\ell \gets \mathbf{H}_\ell + \tilde{\mathbf{X}}_\ell$.
		\ENDFOR
		
		\STATE \textbf{Shared backbone output:}
		\STATE \quad $\hat{\mathbf{Z}} \gets \hat{\mathbf{X}}_{L} \in \mathbb{R}^{N_g \times D_{\mathrm{uni}}}$.
		\STATE \textbf{Output adapter:} map back to granularity-specific space:
		\STATE \quad $\mathbf{Z}^{(g)} \gets \psi^{\mathrm{out}}_g(\hat{\mathbf{Z}}) \in \mathbb{R}^{N_g \times D_g}$.
		
		\RETURN $\mathbf{Z}^{(g)}$.
	\end{algorithmic}
\end{algorithm}

After the coarse stage assigns a granularity label to each input image, the fine stage selects the corresponding granularity-specific encoder. Each granularity branch begins by partitioning the image into non-overlapping patches of size determined by the assigned granularity (coarse: \(16\times16\), medium: \(8\times8\), fine: \(4\times4\)). Each patch is embedded into a feature token using the granularity-specific patch embedding module, producing a sequence whose length varies across granularities. Before entering the shared backbone, all sequences are projected into a unified feature dimension through a granularity-specific input adapter. This linear transformation aligns heterogeneous embedding spaces into a common latent dimension, enabling all granularities to share the same multi-head self-attention blocks without architectural modification. Within the shared backbone, every token undergoes a series of attention and feed-forward transformations, generating a refined representation independent of the original patch size. Upon exiting the shared blocks, the token sequence is mapped back to its granularity-specific feature space using an output adapter. This projection restores granularity-dependent representational capacity, ensuring that each
branch maintains an appropriate inductive bias aligned with its patch resolution. Finally, class prediction is performed by aggregating the adapted token sequence—typically via a class token or global average pooling—followed by a linear classification head to produce the final logits corresponding to the input granularity.

\section{Training Details}
\subsection{Coarse Stage (Complexity Estimator)}

The coarse stage is trained once on Tiny-ImageNet-200, which provides diverse natural images with varying background clutter and texture complexity. All images are resized to $224\times224$ and normalized to $[0,1]$ before feeding into the coarse module. No label information is used at this stage.

For each image $x$, we compute three handcrafted descriptors on the CPU using OpenCV: a Canny edge ratio (spatial domain), a normalized Shannon entropy (intensity statistics), and a high-frequency ratio in the Fourier spectrum (frequency domain). These three values are stacked into a feature vector $\mathbf{s}(x)\in\mathbb{R}^3$ and
combined by a learnable softmax-normalized weight vector $\tilde{\mathbf{w}}$. The resulting scalar is passed through a sigmoid to obtain a complexity score $\Phi(x)\in(0,1)$.

To obtain three granularity regions without supervision, we first compute $\Phi(x)$ for all training images with an initial $\tilde{\mathbf{w}}=[1,1,1]$. The scores are linearly normalized to $[0,1]$, and we then determine two quantiles $q_1,q_2$ such that the dataset is approximately split into three equally sized subsets: $\{\Phi(x)<q_1\}$, $\{q_1\le\Phi(x)<q_2\}$, and $\{\Phi(x)\ge q_2\}$. These fixed intervals define the pseudo granularity labels (coarse / medium / fine), and the corresponding targets $\Phi(x)$ are kept fixed during training.

The complexity estimator is trained by regressing the predicted score $\hat{\Phi}(x)$ towards the fixed target score $\Phi(x)$ using a mean squared error loss. We optimize only the feature weights $\mathbf{w}$ and the two thresholds $(\alpha,\beta)$, which are parameterized in a numerically stable way to enforce $0<\alpha<\beta<1$. We use Adam with learning rate $1\times10^{-3}$, batch size $64$, and train for $15$--$20$ epochs on Tiny-ImageNet. The resulting evolution curves of $\alpha$ and $\beta$ are shown in the main paper. After training, the coarse-stage parameters
$(\tilde{\mathbf{w}},\alpha,\beta)$ are \emph{frozen} and reused for all downstream datasets. For each new dataset, we only recompute $\Phi(x)$ and the corresponding granularity index $g(x)$, without retraining the coarse module.

\subsection{Fine Stage (Granularity-Driven Classification)}
The fine stage operates on standard image classification datasets. For each dataset, we initialize the fine backbone from the ImageNet-1k pretrained Swin-Tiny checkpoint and transplant the patch embedding weights to the three granularities as described in the main paper. During fine-stage training, the coarse module is kept frozen: for each mini-batch, we run the coarse stage once to obtain $\Phi(x)$ and the granularity index $g(x)\in\{1,2,3\}$, then route each image to the corresponding fine branch. No gradient is propagated into the coarse stage, because its feature computation relies on non-differentiable OpenCV operators.

Given the predicted granularity $g(x)$, each sample is tokenized with its corresponding patch size and window configuration, and then passes through the shared Transformer backbone and the granularity-specific input/output adapters. All parameters in the shared backbone and adapters are trained jointly using only the standard cross-entropy loss w.r.t.\ the class labels. In other words, the fine stage is trained exactly like a conventional classifier, except that the mini-batch is dynamically split into three sub-batches and processed by different patch/window settings within the same model.

Unless otherwise specified, we use AdamW with learning rate $1\times10^{-4}$, weight decay $1\times10^{-4}$, batch size $32$, and a cosine-annealing learning rate schedule over $50$ epochs. Standard image classification augmentations are adopted (random resized crop to $224\times224$ and random horizontal flip). No label smoothing or auxiliary losses are introduced. During both training and evaluation, the routing is always driven by the \emph{frozen} coarse module, thus
ensuring that the granularity distribution used for optimization is consistent with the one used at test time.

Different datasets naturally exhibit different complexity distributions. If a dataset happens to concentrate on a single granularity (e.g., most images are judged as fine), the model automatically focuses training on the corresponding branch. This does not affect correctness, because the same routing rule is used at inference. In the supplementary results, we further analyze the empirical granularity histograms and discuss how they relate to the inherent difficulty of each dataset.

\section{Model Details}
\begin{table}[!htbp]
	\centering
	\caption{Parameter settings for the three granularity levels used during inference.}
	\small
	\begin{tabular}{ccccccc}
		\toprule
		{$g$} & {$p$} &  {$w$} & {$layer_1$} & {$layer_2$} & {$layer_3$} & {$layer_4$} \\
		\midrule
		1 & 16 &  28 & 56 & 28 & $\emptyset$ & $\emptyset$\\
		2 & 8 &  14 & 56 & 28 & 14 & $\emptyset$\\
		3 & 4 &  7 & 56 & 28 & 14 & 7\\
		\bottomrule
	\end{tabular}
	\label{tab_8}
\end{table}

This section reports the architectural specifications used in all experiments. As described in the main paper, Grc-ViT contains three granularity levels—coarse ($g{=}1$), medium ($g{=}2$), and fine ($g{=}3$). Each granularity is associated with a specific patch size $p_g$, window size $w_g$, and a granularity-dependent transformer depth $\{L^{(g)}_1, L^{(g)}_2, L^{(g)}_3, L^{(g)}_4\}$, corresponding to the four hierarchical stages of the backbone. Table~\ref{tab_8} summarizes all settings used during inference.

For each granularity, an independent patch embedding layer is used to project $p_g{\times}p_g$ patches into the unified token dimension $D_{\mathrm{uni}}{=}192$. Coarse granularity ($p{=}16$) produces fewer tokens and emphasizes global structure, while fine granularity ($p{=}4$) produces dense token maps that preserve local detail.

Although all granularities share the same attention blocks, each level activates a different subset of stages:
coarse uses only the first two stages, medium activates three stages, and fine activates the full four-level hierarchy. This design preserves granularity-specific inductive biases while maximizing parameter sharing.

The window sizes follow a proportional scaling rule, with $w_g = \{28, 14, 7\}$ corresponding to patch sizes $p_g = \{16, 8, 4\}$, respectively. This ensures that the effective receptive field expands consistently from fine to coarse levels, matching the expected visual granularity.

For shared-stage processing, all tokens are mapped into and out of the shared attention core using the granularity-specific input/output adapters $\psi^{\mathrm{in}}_g$ and $\psi^{\mathrm{out}}_g$, implemented as lightweight linear projections. These adapters harmonize the heterogeneous token spaces produced by different patch sizes.
\section{Supplementary Experiments}
\label{sec_sup}
\subsection{Multi-Granularity Patch Analysis}
\label{sec_patch}
\begin{figure}[!ht]
	\centering
	\small
	\subfloat[]{\includegraphics[width=3.5in]{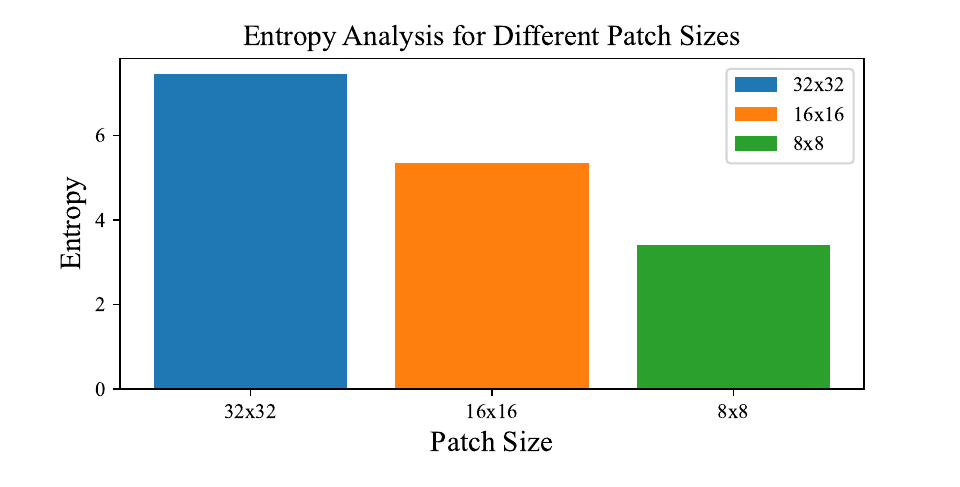}%
		\label{f5-a}}
	\hfil
	\subfloat[]{\includegraphics[width=3.5in]{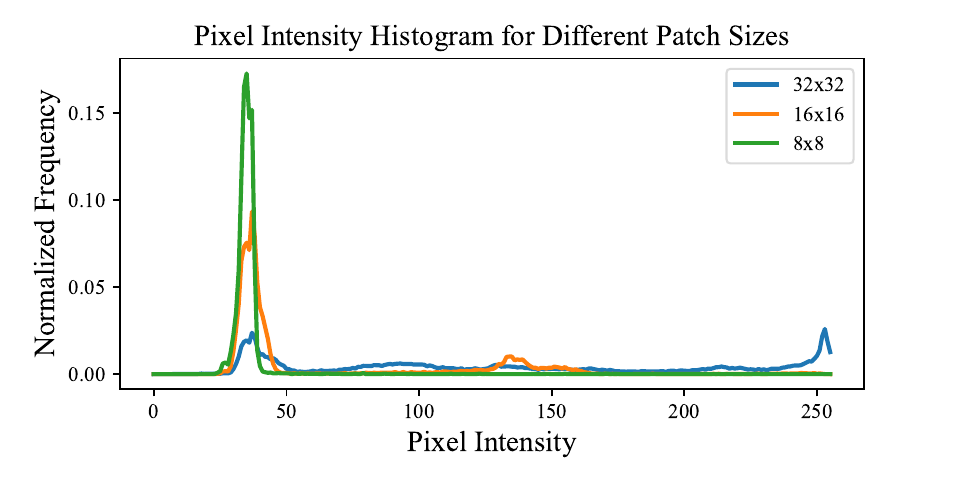}%
		\label{f5-b}}
	\hfil
	\subfloat[]{\includegraphics[width=3.5in]{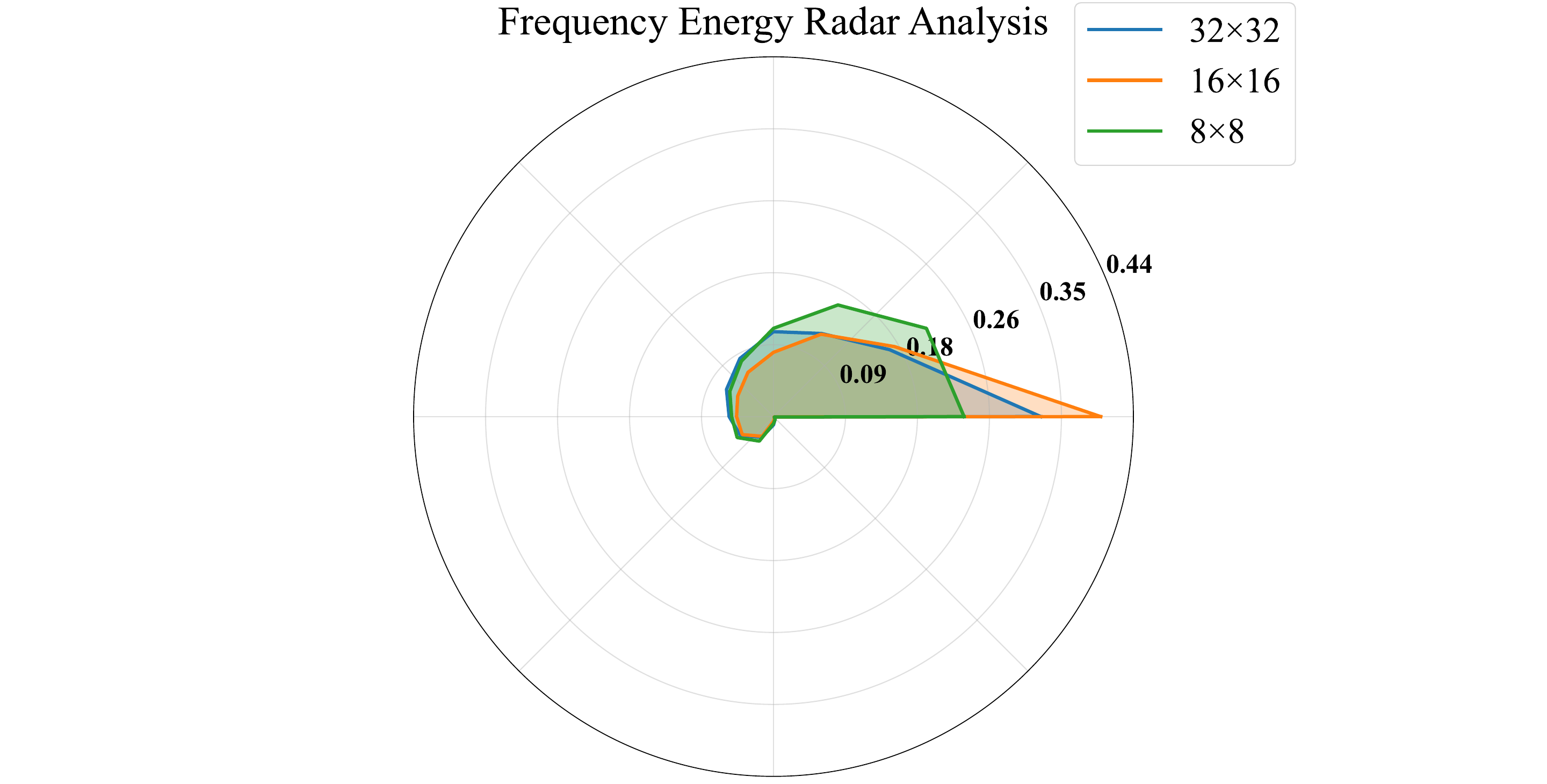}%
		\label{f5-c}}
	\caption{(a) Comparative histogram of image entropy at different Patch\_size values. (b) Line chart of pixel intensity distribution at different Patch\_size values. (c) Radar chart of Fourier transform spectrum at different Patch\_size values.}
	\label{f5}
\end{figure}
Fig.~\ref{f5-a} reports the Shannon entropy of the three patch sizes ($32{\times}32$, $16{\times}16$, $8{\times}8$). We observe that the largest patch ($32{\times}32$) exhibits the highest entropy, while the $8{\times}8$ patch has the lowest value. This indicates that coarse patches tend to mix multiple local structures (texture, edges and backgrounds) within a single region, leading to more diverse gray–level distributions. In contrast, fine patches are more homogeneous and often dominated by a single semantic component (e.g., part of fur or background), which naturally lowers the entropy. This monotonic trend supports our design choice of entropy as one of the complexity descriptors in the coarse stage: patches with richer local patterns are quantitatively distinguished from simpler ones.

Fig.~\ref{f5-b} further compares the normalized pixel–intensity histograms of different patch sizes. The $32{\times}32$ patch shows a broader distribution with heavier tails, reflecting strong contrast between bright and dark regions coexisting within a single receptive field. As the patch size shrinks, the histograms become more concentrated and peak around narrower ranges of intensities, which corresponds to more locally-consistent textures. This behaviour reveals that coarse granularity captures global contrast and large-scale illumination variations, whereas fine granularity focuses on subtle variations inside a relatively uniform area. Such complementary pixel-level statistics are consistent with the granularity intuition of our method.

Fig.~\ref{f5-c} presents a radar chart of the radial frequency energy distribution obtained from 2D FFT. For each patch size, we integrate the spectrum over concentric annuli and normalize the energy across frequency bands. Coarser patches ($32{\times}32$) allocate relatively more energy to low and mid frequency bands, which are associated with smooth structures and large-scale shape transitions. Smaller patches ($16{\times}16$ and $8{\times}8$) exhibit relatively stronger responses in higher-frequency bands, corresponding to sharper edges and fine textures. Although the exact profiles differ from the entropy and histogram plots, all three views consistently indicate that different patch sizes favour different frequency ranges. This validates the use of FFT-based descriptors as a complementary complexity measure and provides a frequency-domain justification for multi-granularity modelling in Grc-ViT.

\subsection{Dataset Complexity Profiling}

We first apply the trained coarse-stage complexity estimator to six commonly used benchmarks and report statistics of the complexity score $\Phi(x)$ as well as the implied granularity distribution. Table~\ref{tab:complexity_stats} lists the granularity composition of the datasets. It shows that simpler datasets like CIFAR-10/100 are dominated by coarse-grained samples. Conversely, fine-grained datasets such as Stanford Dogs, Aircraft, and CUB contain a substantially higher proportion of complex, fine-grained samples. The distributions closely match human intuition and confirm that the coarse stage provides meaningful data-driven routing signals.

\begin{table}[!htbp]
	\centering
	\caption{Statistics of the complexity score $\Phi(x)$ and granularity distribution for six datasets. }
	\small
	\begin{tabular}{lcccc}
		\toprule
		Dataset & Mean   &
		Coarse(\%) & Medium(\%) & Fine(\%) \\
		\midrule
		CIFAR-10        & 0.21   & 74.1 & 23.7 & 2.2 \\
		CIFAR-100       & 0.28  & 32.8 & 58.3 & 8.9 \\
		Tiny-ImageNet   & 0.41   & 33.4 & 22.1 & 44.5 \\
		FGVC Aircraft   & 0.63   &  7.5 & 28.2 & 64.3 \\
		Stanford Dogs   & 0.67   &  3.2 & 20.5 & 76.3 \\
		CUB-200-2011    & 0.71   &  6.8 & 17.6 & 75.6 \\
		\bottomrule
	\end{tabular}
	\label{tab:complexity_stats}
\end{table}

The complexity distributions reveal a natural ordering of dataset difficulty. Low-resolution datasets (CIFAR-10/100) contain fewer edges and less structural variation, yielding predominantly coarse assignments. Tiny-ImageNet exhibits moderate complexity due to diverse backgrounds. Fine-grained benchmarks require discriminating subtle part-level differences, thus receiving higher $\Phi(x)$ and significantly more fine-granularity routing. These results validate the coarse-stage estimator as both interpretable and aligned with human visual intuition.

\subsection{Ablation on Routing Strategy}

We evaluate how different routing strategies affect classification performance:
(a) using the fine branch exclusively (equivalent to a Swin baseline),
(b) forcing all images to use the same granularity $g$,
(c) random routing, and
(d) our complexity-driven granularity selection.
Table~\ref{tab:routing_ablation} shows that removing dynamic routing significantly degrades performance. Random routing performs worst due to mismatching granularity and image structure. Our method consistently achieves the best accuracy.

\begin{table}[!htpp]
	\centering
	\caption{Routing ablation.}
	\small
	\begin{tabular}{lccc}
		\toprule
		Method & Dynamic & Granularity & Acc(\%) \\
		\midrule
		Fixed fine only   & $\times$  & fine       & 80.5 \\
		Fixed $g$       & $\times$  & fixed      & 76.1 \\
		Random routing     & $\times$  & random     & 72.3 \\
		\textbf{Grc-ViT (ours)} & \checkmark  & adaptive    & \textbf{81.6} \\
		\bottomrule
	\end{tabular}
	\label{tab:routing_ablation}
\end{table}

The experiments confirm that the routing mechanism is essential. Fixed granularity prevents the model from adapting to varying image complexity. Random routing disrupts the alignment between visual structure and patch resolution. Our coarse-stage routing establishes the correct granularity per sample and delivers the highest accuracy.

\subsection{Ablation on Backbone Sharing}

We further compare sharing one unified backbone (with adapters) against training three independent Swin backbones. Table~\ref{tab:backbone_ablation} indicates the three-backbone variant uses more parameters. In contrast, the proposed shared backbone plus adapters significantly reduces the parameter count, indicating that multi-granularity inductive biases can be preserved with much higher efficiency.

\begin{table}[!htbp]
	\centering
	\caption{Backbone sharing ablation. }
	\small
	\begin{tabular}{lccc}
		\toprule
		Model Variant & Backbones & Params(M) & FLOPs(G) \\
		\midrule
		3× independent Swin     & 3 & 87.4 & 13.42  \\
		\textbf{Grc-ViT (shared)} & 1 & \textbf{61.7} & \textbf{6.07}  \\
		\bottomrule
	\end{tabular}
	
	\label{tab:backbone_ablation}
\end{table}

Training three independent networks causes severe parameter fragmentation and prevents effective cross-granularity feature reuse. Grc-ViT, in contrast, shares a unified attention core while using lightweight input/output adapters to preserve granularity-specific inductive biases. This design yields both efficiency and accuracy gains over non-sharing architectures.

 \end{document}